\documentclass[10pt, a4paper]{article}
\usepackage[]{lrec-coling2024} 
\usepackage{mathrsfs}
\usepackage{amsmath}
\usepackage{hyperref}
\usepackage{times}
\usepackage{latexsym}
\usepackage{booktabs}
\usepackage{multirow}
\usepackage[T1]{fontenc}

\usepackage[utf8]{inputenc}
\usepackage{multirow}

\usepackage{microtype}
\usepackage{graphicx}
\usepackage{inconsolata}

\usepackage{subcaption}
\title{FCDS: Fusing Constituency and Dependency Syntax into Document-Level Relation Extraction}

\name{Xudong Zhu, Zhao Kang$^{\ast}$ \thanks{*Corresponding author}, Bei Hui}

\address{
        University of Electronic Science and Technology of China,
        Sichuan, China \\
        \{2020080902004, zkang, bhui\}@uestc.edu.cn\\
     }

\abstract{
Document-level Relation Extraction (DocRE) aims to identify relation labels between entities within a single document. It requires handling several sentences and reasoning over them. State-of-the-art DocRE methods use a graph structure to connect entities across the document to capture dependency syntax information. However, this is insufficient to fully exploit the rich syntax information in the document. In this work, we propose to fuse constituency and dependency syntax into DocRE. It uses constituency syntax to aggregate the whole sentence information and select the instructive sentences for the pairs of targets. It exploits the dependency syntax in a graph structure with constituency syntax enhancement and chooses the path between entity pairs based on the dependency graph. The experimental results on datasets from various domains demonstrate the effectiveness of the proposed method. The code is publicly available at \href{https://github.com/xzAscC/FCDS}{https://github.com/xzAscC/FCDS}.
 \\ \newline  }

\begin{document}

\maketitleabstract

\section{Introduction}
Relation Extraction (RE) is a crucial task in information extraction that aims to model relational patterns between entities in an unstructured text. There are two specific scenarios: sentence-level RE and document-level RE. Unlike sentence-level RE \cite{sentenceRE-Dixit, sentenceRE-Lyu}, where entities are located in the same sentence, document-level RE (DocRE) identifies the relation labels between entities within a document. Therefore, DocRE better meets practical needs and has recently received increasing attention \cite{zhou2021document, Zhao2022DocumentlevelRE}. \par

\begin{figure}
\centering{\includegraphics[height=0.57\linewidth]{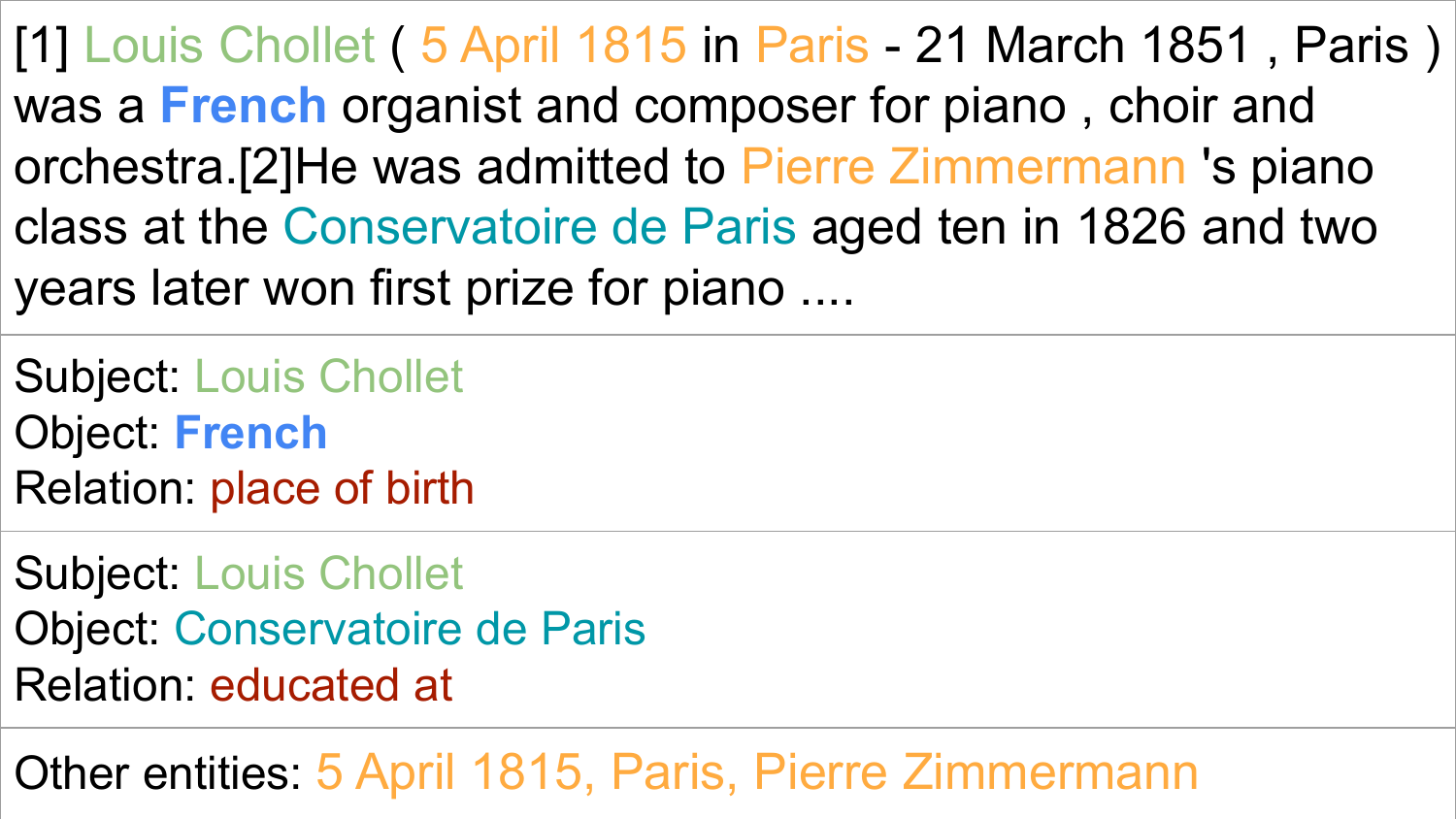}}
\caption{A sentence-level and a document-level relation instance from DocRED. Entity pairs are colored differently according to relation. To identify irrelevant or even noisy relations, unrelated entities are uniformly labeled with one single \textcolor[RGB]{115,65,65}{\emph{color}}.
}
\label{fig_exam}
\end{figure}

A formidable obstacle confronting DocRE is inferring relations of entity pairs in long sentences, which often contain irrelevant or even noisy information \cite{gupta2019neural}. Figure \ref{fig_exam} is an example, which includes a sentence-level relation and a document-level relation from DocRED. 
 To infer the relation between Louis Chollet and Conservatoire de Paris, models should exclude the influence of unrelated entities and figure out that the word `He' in sentence[2] refers to `Louis Chollet'. Buried under massive irrelevant information, DocRE models often struggle with intricate relation instances. Therefore, implicitly learning an instructive context is not sufficient for DocRE \cite{bai-etal-2021-syntax}. \par

In a document, interactions between entities are complex. Pre-trained language models (PLMs) \cite{BERT} have shown great potential in many downstream tasks. Some work \cite{ye-etal-2020-coreferential, zhou2021document} implicitly captures such interactions between entities through PLMs. Others, however, model this information explicitly. They first construct document graphs \cite{GAIN, liu2023document} that consist of different nodes (e.g., mentions, entities, sentences, or the document) to turn instructive context into graphs. Since syntax information can help DocRE by providing explicit syntax refinement and subsentence modeling \cite{duan-etal-2022-just}, recent studies \cite{sahu2019inter, SagDRE} adopt a dependency graph to incorporate both syntax information and structural context. They find that a structural graph can better capture relations and shorten the distance between entities. However, as pointed out in \cite{sundararaman2019syntax, bai-etal-2021-syntax}, although PLMs are trained with massive real-world text data, there is still a great gap between the implicitly learned syntax and the golden syntax. In fact, syntax information is widely used in sentence-level RE  \cite{xu-etal-2016-improved, qin-etal-2021-relation}, but it has not yet been fully explored under the DocRE scenario.\par

\begin{figure}[t]
\begin{subfigure}{1.0\columnwidth}
\includegraphics[width=\textwidth]{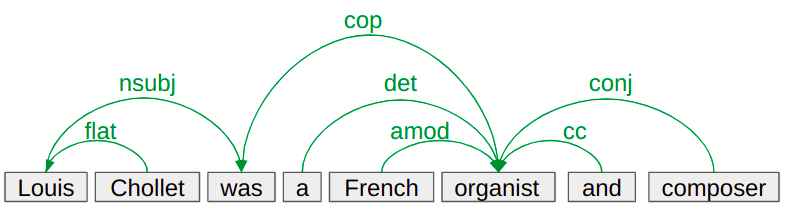}
\caption{Dependency tree describes dependencies between words within a single sentence. Exploiting such syntax information can significantly complement the original plain text and capture the couplings among neighbors. Our model converts the dependency tree to a dependency graph.
} 
\label{fig_syn_a}
\vspace{0.5cm}
\end{subfigure}
\begin{subfigure}{1.0\columnwidth} 
\includegraphics[width=\textwidth]{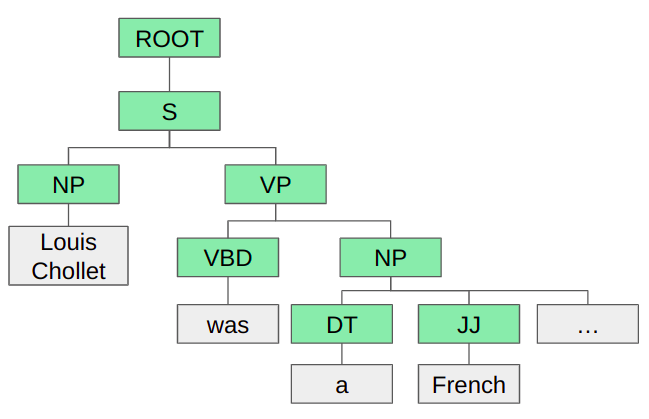}

\caption{Constituency tree organizes the sentence in a tree structure, which not only induces extra hierarchical syntax information but also enables exploring subsentences in arbitrary granularity. Non-leaf nodes are colored green. }
\label{fig_syn_b}
\end{subfigure}
\caption{Syntactic parsing results of evidence sentence ``\emph{Louis Chollet ...}” mentioned in the previous relation instance. (a) and (b) represent the corresponding dependency and constituency tree, respectively.}
\end{figure}

To fully exploit the syntax information in the document, we fuse constituency and dependency syntaxes in this paper. We mainly adopt dependency graphs and constituency trees to incorporate extra syntax information, and use information from the constituency tree to further enhance the representation of the dependency graph. The dependency and constituency syntax depict complementary but different aspects of syntax information. The dependency graph in figure \ref{fig_syn_a} is mainly used to integrate syntactic information within a single sentence that strongly complements the original plain text, while the constituency tree in figure \ref{fig_syn_b} organizes different words of a single sentence hierarchically and reasonably. 

We observe that dependency syntax is better for constructing paths between entity pairs \cite{SagDRE}, while constituency syntax is better for aggregating sentence-level information. Therefore, we follow previous studies by transforming the dependency tree into a graph and extracting the paths between entity pairs. However, the gap between the learned syntax in PLMs and the golden syntax for dependency trees has heavily influenced the performance of DocRE. To address this issue, we propose to utilize the constituency tree to aggregate sentence-level information to compensate for the gap in the dependency tree. Specifically, we utilize a single-layer MLP to fuse the sentence root in the constituency tree and the dependency graph to replace the original sentence root in the dependency graph. Furthermore, in order to better consider the sentence interaction in the dependency graph, we add a document-level node and link each sentence root to reduce the distance of entity pairs and better capture long-distance relations.
Through extensive experiments on three public DocRE benchmarks, DocRED \cite{DOCRED}, CDR \cite{li2016biocreative}, and GDA \cite{GDA}, we demonstrate that our model outperforms existing methods.\par

Our key contributions in this work can be summarized as follows:
\begin{enumerate}
    \item We propose to utilize the constituency tree to aggregate sentence-level information to compensate for the gap in the dependency tree and improve the dependency graph by adding a document node to reduce the distance of entity pairs and simplify long-sentence interaction.
    \item We process the dependency graph and the constituency tree with Tree-LSTM and GCN, respectively, and set a learnable parameter to adjust their weights.
    \item The results of the experiments demonstrate that our model outperforms the existing methods on three DocRE benchmarks, especially on DocRED, where our model improves the IgnF1 of the state-of-the-art methods by at least 1.56\%.
\end{enumerate}

\section{Relation Works}
\subsection{Document-level Relation Extraction}
DocRE is a challenging task since long sentence learning usually requires effective long-distance feature encoding and reasoning \cite{sahu2019inter}. To tackle this challenge, some methods apply PLMs for more informative contextual token encoding. \cite{Tang2020HINHI} proposes a hierarchical inference network from the level of entities, sentences, and documents using BERT, while \cite{ye-etal-2020-coreferential} explicitly encodes the coreference information to improve the coreferential reasoning ability of BERT. \ \cite{xie2022eider} empowers DocRE by efficiently extracting evidence and effectively integrating the extracted evidence in inference.\par

In addition to BERT-based methods, another line of research proposes to use the graph structure to shorten the distances between entities in the document. \cite{GAIN} uses two graphs to represent mention-level and entity-level relations, respectively, while \cite{SagDRE} employs linguistic tools to build various edges, such as coreference edges, which embed inter-sentence and intra-sentence dependencies. \cite{Xu2020DocumentLevelRE} enforces the model to reconstruct reasoning paths while identifying correct relations. \cite{duan-etal-2022-just} utilizes the constituency tree to obtain evidence for DocRE and incorporates dependency graphs to classify the relations. \par

However, these methods either use regular graph structures that cannot capture sequential information in the original text \cite{GAIN, Xu2020DocumentLevelRE}, or use dependency and constituency information separately \cite{SagDRE, duan-etal-2022-just}. This work overcomes this drawback by incorporating both constituency and dependency information and enhancing the dependency graph with a constituency tree. With different syntax incorporated, our model can fuse dual-granularity information and better capture long-distance relations.



\subsection{Constituency and Dependency Syntax}

Since syntax intuitively shares many common features with RE, syntactic features are a highly effective DocRE performance enhancer, according to several empirical verifications in previous work \cite{GAIN, SagDRE, xie2022eider}. In particular, dependency syntax is extensively studied in DocRE, while constituency syntax is overlooked.\par

Although the constituency and dependency syntaxes share some common syntactic information, they characterize it from different perspectives. Some work has revealed the mutual benefits of integrating these two heterogeneous syntactic representations for various NLP tasks. \cite{zhou-zhao-2019-head, strzyz-etal-2019-sequence} integrate dependency and constituency syntactic information as a representation of a parse tree or sequence. \cite{fei-etal-2021-better}, which is designed for the Semantic Role Labeling (SRL) task, converts the dependency and constituency trees into graphs and performs the graph learning strategy on them. \cite{Dong2022SyntacticML} proposes to map phrase-level relations in the constituency tree into word-level relations and adopts multi-view learning to capture multiple relationships from the constituency graph and dependency graph for the Open Information Extraction (OpenIE) task, which is the most relevant model to ours.\par

Our model differs from \cite{Dong2022SyntacticML} mainly in two aspects: \textbf{(1)} \cite{Dong2022SyntacticML} turns the constituency tree into a graph by heuristic rules and aligns instances of the same node across the dependency graph and constituency graph. Our model, however, utilizes Tree-LSTM to handle the constituency syntax in tree form and compensate for the inaccuracy of dependency; \textbf{(2)} To link different syntax, we utilize the constituency tree to enhance the dependency graph instead of adopting multi-view learning to fuse heterogeneous information from both graphs.\par

\section{Methodology}
A document $\mathcal{D}$ contains $I$ sentences $\{sen_i\}_{i=1}^{I}$ and $N$ entities $\{e_i\}_{i=1}^{N}$. $sen_i$ is the $i^{th}$ sentence, which includes $P_i$ tokens: $\{t_{i,1},t_{i,2},\cdots, t_{i,P_i}\}$. An entity $e_k$ can have $Q_k$ mentions $\{m_{k,1},m_{k,2},\dots, m_{k,Q_k}\}$. \par
The goal of DocRE is to correctly infer all relations between each entity pair $(e_s,e_o)_{s,o=1,2,\cdots, N;s\neq o}$, where $e_s$ is a subject entity and $e_o$ is an object entity. The predicted relations are subsets of the predefined relation set $\mathcal{R}$ or $\{\mathcal{NA}\}$ (without relation). \par
\begin{figure*}[!t]
\centering{\includegraphics[height=0.5\linewidth]{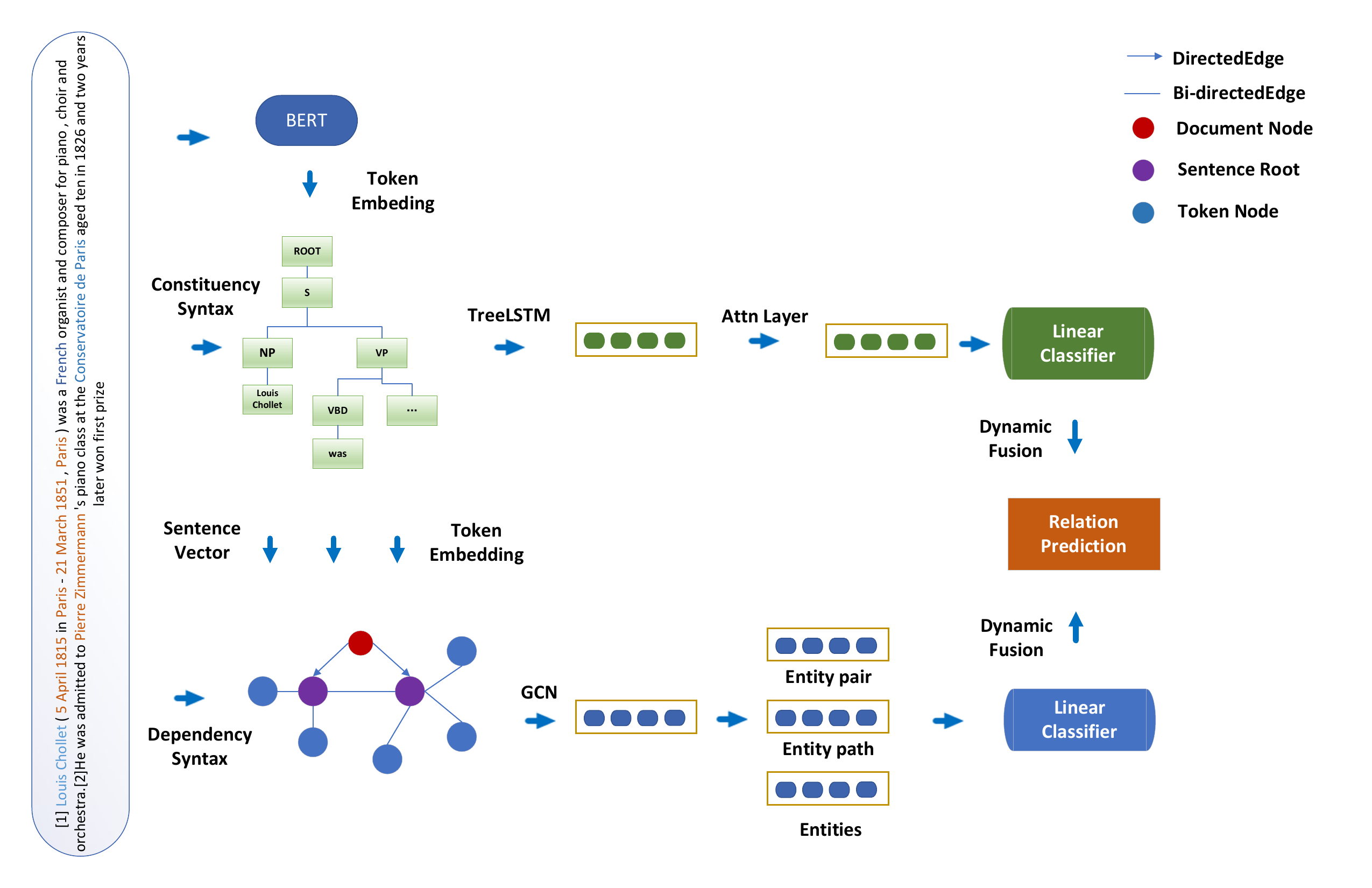}}
\caption{The overview of our architecture. Note that we use the result of constituency syntax to enhance the dependency graph and obtain relations between entity pairs with dynamic weighted fusion.}
\label{overview}
\end{figure*}

The overall architecture of the Fusing Constituency and Dependency Syntax (FCDS) is illustrated in Figure \ref{overview}. We exploit dependency and constituency syntax to build a dependency graph and constituency tree and utilize BERT to encode words in the document. Then we use Tree-LSTM to aggregate information from the constituency tree and exploit the dependency graph by graph neural network (GNN) \cite{GCN, DAI2022107659} while utilizing the constituency tree to improve the dependency graph. We obtain relations between entity pairs by a learnable weight to combine the dependency graph and the constituency tree. \par


\subsection{Text Encoding}\label{text encoding}

Given a document, a special marker `*' \cite{zhang-etal-2017-position} will be first inserted before and after each mention. Then we feed tokens from document $\mathcal{D}$ into PLMs to obtain contextualized representation $\mathcal{H}=\{h_1, \dots, h_T\} \in \mathcal{R}^{T \times d}$, where $T$ is the number of tokens and $d$ is the dimension of token embedding:
\begin{equation}\label{PLM}
    \mathcal{H} = PLM \{x_{1,1},\cdots, x_{I,P_I}\}
\end{equation}
 $x_{i,j}$ is the $j^{th}$ word for the $i^{th}$ sentence in the document and the PLMs can be a pre-trained BERT \cite{BERT} or LSTM model.

\subsection{Constituency Tree Construction}\label{constituency tree construction}
After obtaining the token-level embedding through the PLMs model, FCDS utilizes a constituency tree to compensate for the inaccuracy in the dependency tree based on hierarchical syntax information.
Specifically, we take advantage of the constituency tree based on the document $\mathcal{D}$ and use Tree-LSTM \cite{miwa-bansal-2016-end, duan-etal-2022-just} to incorporate root information of a sentence. First, we parse each sentence into the corresponding constituency tree\footnote{constituency trees are obtained using the Stanza library. \href{https://stanfordnlp.github.io/stanza/} { https://stanfordnlp.github.io/stanza/}}. We can see from Figure \ref{fig_syn_b} that each constituency tree describes a logical way to restore the entire sentence piece by piece. Using a constituency tree, we can not only incorporate extra hierarchical syntax information but also encode sentences with arbitrary granularity.\par
To represent the constituency tree with Tree-LSTM, we first initialize hidden state $h_j$ and memory cell state $c_j$ with zeros. The input vector of leaf nodes is initialized with their corresponding representations inside PLMs, while non-leaf nodes are set to zeros. We then broadcast features of leaf nodes all the way up to the root node using Tree-LSTM \cite{miwa-bansal-2016-end}. The input gate $i_j$, the output gate $o_j$, and the forget gate $f_{jk}$ of an arbitrary node $j$ in the constituency tree are calculated as: 
\begin{equation}
     i_j = \sigma(W_ix_j +\sum_{l \in Child(j)}h_{l}W_{il}+b_i) 
\end{equation}
\begin{equation}
     o_j = \sigma(W_ox_j +\sum_{l \in Child(j)}h_{l}W_{ol}+b_o) 
\end{equation}
\begin{equation}
     f_{jk} = \sigma(W_fx_j +\sum_{l \in Child(k)}h_{jl}W_{kl}+b_f)
\end{equation}
where $\sigma$ is the sigmoid function, $W, b$ are trainable parameters. $x_j$ denotes the input vector of node $j$ and $Child(j)$ means the child of node $j$ in the constituency tree.\par
The integrated result $u_j$ is calculated as:
\begin{equation}
     u_j = \tanh(W_ux_j +\sum_{l \in Child(j)}h_{l}W_{ul}+b_u)
\end{equation}
\par At last, we update hidden state $h_j$ and memory cell state $c_j$ as follow,
\begin{equation}
     c_j = i_j \odot u_j + \sum_{l \in Child(j)} f_{jl} \odot c_l
\end{equation}
\begin{equation}
     h_j = o_j \odot \tanh(c_j)
\end{equation}

\par In practice, we extract information from its child nodes recurrently and use the sentence root feature as the sentence feature vector. For a document with $I$ sentences, we can get all the sentence vectors $\{s_i\}_{i=1}^{I}$. Since not all sentences contain relevant information for relation reasoning, we employ a multi-head attention layer over the sentence vector to identify the most relevant ones. We formulate this process as:
\begin{equation}
    V_{doc} = [s_1, s_2, \dots, s_I]
\end{equation}

\begin{equation}\label{Attn}
    S,\  A = Attn(W_{t1}(e_s-e_o), W_{t2} V_{doc}, W_{t3} V_{doc})
\end{equation}
where $W_{t1}, W_{t2}$, and $W_{t3}$ are trainable weights and $Attn$ represents an attention layer. $A$ is the attention score and $S$ is the sentence vector through the attention layer. Then, we combine the weighted sentence with entity pairs and calculate the score $z_{const}$ of relation $r$:
\begin{equation}
    z_s = tanh(W_{(s1)}e_s+W_{(s2)}S)
\end{equation}
\\
\begin{equation}
    z_o = tanh(W_{(o1)}e_o+W_{(o2)}S)
\end{equation}
\\
\begin{equation}
    z_{const} = z_s^\top W_{s, o}z_o+b_{const}
\end{equation}
where $W, b$ are trainable parameters. As a result, we obtain the relation scores $z_{const}$ based on the constituency tree with attention score $A$ and sentence vector $S$ to enhance the dependency graph. \par

\subsection{Dependency Graph Construction}\label{dependency graph construction}
After obtaining the above representation, we construct a dependency graph to aggregate information of syntactically associated words. Each sentence in the document is fed into a dependency parser, which generates a dependency syntax tree. Then we convert the dependency trees to dependency graphs. Note that we add a document node to shorten the distance between entity pairs and better aggregate information of long-distance sentences. \par

The dependency graph contains four kinds of nodes: non-root token nodes, root token nodes, mention nodes, and document nodes. Specifically, each token in the document corresponds to a token node, and for tokens that are not the sentence root in the dependency tree, its encoded feature corresponds to its node feature. 
For sentence root, we treat it especially to combat possible error parsing, taking advantage of the root feature $S$ obtained by the Eq. \ref{Attn}. We use MLP to fuse the original feature with $S$ in Eq. \ref{Attn} since the sentence root in the dependency graph is expected to include information of the entire sentence. For the mention node, the node feature is calculated by averaging the features of tokens in this mention. The document node, as a node that includes information of each sentence root and emphasizes the conducive sentences, is calculated as the weighted average of each sentence vector. The weight is the attention score of the entity pairs in Eq. \ref{Attn} . \par

There are four types of edges in this graph. Three of them are bi-directed, and one is directed. Bi-directed dependency syntax edges are added between each pair of connected tokens in the syntax tree. Then the bi-directed edges are added between the dependency syntax tree roots of adjacent sentences, since there exist close context relationships between adjacent sentences. As each sentence in the document serves the same topic of this document, bi-directed edges are added between dependency syntax tree roots and the document node. The last type of edge is directed and exists between non-adjacent sentence roots to capture long-distance information and embed sequential information. \par

The weight of bi-directed edges is 1 to inform their strong connection. The weight $ADJ_{i,j}$ of directed edges from non-adjacent sentence root node $i$ to $j$ are calculated based on their feature vectors:
\begin{equation}\label{edge weight}
    ADJ_{i,j} = \frac{S_i \cdot S_j}{||S_i|| \cdot ||S_j||}
\end{equation}
where $S_i$ and $S_j$ are the sentence vector $i$ and $j$ in Eq. \ref{Attn}. Using these learned weights, our model can obtain a logic flow from the earlier root to the later root automatically and fuse the information between dependency and constituency syntax to aggregate information from different perspectives. \par

After obtaining the adjacent matrix $ADJ$ of the dependency graph, we employ GCN for feature aggregation and entity strengthening.
\begin{equation}
    q^{l+1} = GCN(q^{l}, ADJ)
\end{equation}
where $q^l$ is the input feature of layer $l$ and $q^{l+1}$ is the output feature.\par
Then entity representation $entity_i$ is abstracted by merging the embeddings of all mentions of this entity based on logsumexp \cite{jia-etal-2019-document}:
\begin{equation}\label{logsumexp}
    entity_i =  log \sum_{j} exp(m_{i,j})
\end{equation}
where $m_{i,j}$ is $j^{th}$ mention embedding of $entity_i$.\par
In addition to entity-specific embeddings, we also extract the shortest path \ $path_{s,o}$ between two targeted entities to complement the entity pair $(e_s, e_o)$ on the dependency graph.
\begin{equation}
    path_{s,o} = [ e_s, node_1, node_2, \cdots, node_n, e_o ]
\end{equation}
For efficiency, we limit the maximum length of the selected path to 14, which means that a maximum of 12 path nodes will be selected except the head and tail entities. Any path longer than 12 will select the first 12 nodes, while a path less than 12 will be filled with zero tensor.  In addition, we use MLP to explore the relation of entity pairs.
\begin{equation}
    pair_{s,o} = LeakyRelu(W_{p1}e_s + W_{p2}e_o)
\end{equation}
\par Through the above steps, dependency syntax complemented entity and context representations are acquired. Following the previous methods \cite{mou-etal-2016-natural}, we then concatenate them all to strengthen the features of this entity pair:
\begin{equation}
    I_{s,o} = [e_s;e_o;pair_{s,o};path_{s,o}]
\end{equation}
We compute the relation score $z_{dep}$ based on dependency graph:
\begin{equation}
    z_{dep} = W_{(d2)} \sigma (W_{(d1)}I_{s,o} + b_{(d1)}) + b_{(d2)}
\end{equation}
where $W, b$ are trainable parameters.\par

\subsection{Dynamic Fusion and Classification}\label{relation head prediction}
Finally, we combine two scores acquired by the dependency graph and constituency tree as a dynamic weighted sum of them \cite{kendall2018multi}.
\begin{equation}\label{eta}
    z_{final} = z_{dep} + \eta z_{const}
\end{equation}
\par We adopt adaptive margin loss as loss function \cite{SagDRE}.
\begin{equation}\label{loss}
    \mathcal{L} = \sum_{1 \le i \le C} max(0, \alpha-c_i(z^i_{final} - z^s_{final}))
\end{equation}
where $\alpha > 0$ is a hyper-parameter for margin and $c_i$ is 1 if the sample belongs to the positive class and -1 otherwise. C is the number of classes and $z^i_{final},\  z^s_{final}$ is the final score of \ $non\mathcal{NA}, \mathcal{NA}$ classes. Note that the proposed adaptive margin loss is reduced to Hinge loss \cite{Gentile1998LinearHL} in the binary RE tasks.

\section{Experiments}

\subsection{Datasets}
To comprehensively evaluate our model, we assess the proposed model on three document-level datasets from various domains. Statistics of these datasets are listed in Table \ref{datasets}.

\begin{itemize}
    \item \textbf{DocRED} \cite{DOCRED} is a large-scale human-annotated dataset constructed from Wikipedia and Wikidata. It contains 132,275 entities, 56,354 relational facts and 96 relation classes. More than 40.7\% of the relation pairs are cross-sentence relation facts.
    \item \textbf{CDR} \cite{li2016biocreative} is a biomedical DocRE dataset built from 1,500 PubMed abstracts that are randomized into three equal parts for training, validation and testing. The task is to predict the binary relation between Chemicals and Diseases.
    \item \textbf{GDA} \cite{GDA} is also a biomedical DocRE dataset contains 30,192 abstracts. The dataset is annotated with binary relations between Gene and Disease concepts using distant supervision. 
\end{itemize}

\subsection{Implementation details}
Our model is implemented on Pytorch \cite{paszke2019pytorch} and uses stanza \cite{qi2020stanza} to extract constituency and dependency syntax. For all experiments, the learning rate is set to 5e-5 and the weight decay is 1e-4. The GCN layer number for the dependency graph is set to 3 and the output dimension is 128. The hidden state and cell state of each node in the constituency tree share a dimension of 256. $\alpha$ in Eq. \ref{loss} is set to 1.0 and learning rate warmup \cite{DBLP:journals/corr/GoyalDGNWKTJH17} with ratio 0.06 is implemented followed by a linear decay to 0. The entire model is optimized by AdamW \cite{loshchilov2018decoupled}.
    
\subsection{Results on DocRED}

We compare our model with graph-based methods and BERT-based methods in DocRED. For BERT-based methods, we compare the proposed method with BERT \cite{DOCRED}, ATLOP \cite{zhou2021document}, evidence-based EIDER \cite{xie2022eider}, and self-training method DREEAM 
 \cite{ma-etal-2023-dreeam}. Graph-based models include LSR \cite{nan-etal-2020-reasoning}, HeterGSAN \cite{Xu2020DocumentLevelRE}, DRE \cite{xu-etal-2021-discriminative}, CorefDRE \cite{xue2022corefdre}, GAIN \cite{GAIN}, SagDRE \cite{SagDRE}, and LARSON \cite{duan-etal-2022-just}. Following previous work \cite{zhou2021document}, we train our model on $\mathrm{BERT}_{base}$ and $\mathrm{DeBERTa}_{Large}$. We report not only F1 and Ign F1 (F1 score excluding the relational facts shared by the training and dev/test set) following the prior studies \cite{DOCRED}, but also Intra F1 (F1 that only considers intra-sentence relational facts) and Inter F1 (F1 that only considers inter-sentence relational facts). \par
\begin{table}
\caption{Statistics of three benchmarks used in our experiments.}
\label{datasets}
\resizebox{\linewidth}{!}{
\begin{tabular}{cccc}
\hline
Statistics               & DocRED & CDR & GDA   \\ \hline
\# Train                 & 3053   & 500 & 23353 \\
\# Dev                   & 1000   & 500 & 5839  \\
\# Test                  & 1000   & 500 & 1000  \\
\# Relations             & 96     & 2   & 2     \\
Avg.\# sentences per Doc & 8.0    & 9.7 & 10.2  \\ \hline
\end{tabular}}
\end{table} 

\begin{table*}
\caption{Results (\%) of relation extraction on the dev and test set of DocRED. Results of other methods are directly taken from original papers.}
\label{results in docred}
\begin{tabular}{@{}llllcccccc@{}}
\toprule
\multicolumn{4}{c}{\multirow{2}{*}{\textbf{Model}}} & \multicolumn{4}{c}{\textbf{dev}}             & \multicolumn{2}{c}{\textbf{test}} \\ \cmidrule(l){5-10} 
\multicolumn{4}{c}{}                       & \textbf{IgnF1} & \textbf{F1}    & \textbf{Intra F1} & \textbf{Inter F1} & \textbf{IgnF1}       & \textbf{F1}         \\ \midrule
\multicolumn{4}{l}{$\mathrm{BERT}_{base}$\cite{DOCRED}}                   & -     & 54.16 & 61.61    & 47.15    & -           & 53.20      \\

\multicolumn{4}{l}{$\mathrm{LSR-BERT}_{base}$\cite{nan-etal-2020-reasoning}}                   & 52.43 & 59.00 & 65.26   & 52.05   & 56.97       & 59.05   \\
\multicolumn{4}{l}{$\mathrm{GAIN-BERT}_{base}$\cite{GAIN}}                   & 59.14 & 61.22 & 67.10    & 53.90    & 59.00       & 61.24      \\
\multicolumn{4}{l}{$\mathrm{HeterGSAN-BERT}_{base}$\cite{Xu2020DocumentLevelRE}}                   & 58.13 & 60.18 & -   & -   & 57.12       & 59.45   \\
\multicolumn{4}{l}{$\mathrm{DRN-BERT}_{base}$\cite{xu-etal-2021-discriminative}}                   & 59.33 & 61.09 & -   & -   & 59.15       & 61.37   \\
\multicolumn{4}{l}{$\mathrm{ATLOP-BERT}_{base}$\cite{zhou2021document}}                  & 59.22 & 61.09 & -        & -        & 59.31       & 61.30      \\
\multicolumn{4}{l}{$\mathrm{CorefDRE-BERT}_{base}$\cite{xue2022corefdre}}                  & 60.85 & 63.06 & -        & -        & 60.78       & 60.82      \\
\multicolumn{4}{l}{$\mathrm{EIDER-BERT}_{base}$\cite{xie2022eider}}                  & 60.51 & 62.48 & 68.47    & 55.21    & 60.42       & 62.47      \\
\multicolumn{4}{l}{$\mathrm{SagDRE-BERT}_{base}$\cite{SagDRE}}                 & 60.32 & 62.06 & -        & -        & 60.11       & 62.32      \\
\multicolumn{4}{l}{$\mathrm{LARSON-BERT}_{base}$\cite{duan-etal-2022-just}}                 & 61.05 & 63.01 & 68.63        & 55.75        & 60.71       & 62.83      \\
\multicolumn{4}{l}{$\mathrm{DREEAM-BERT}_{base}$\cite{ma-etal-2023-dreeam}}                 & 60.51 & 62.55 & -        & -        & 60.03       & 62.49      \\
\multicolumn{4}{l}{$\mathrm{\textbf{FCDS-BERT}}_{base}$}              & \textbf{62.61} & \textbf{64.42} & \textbf{68.79}    & \textbf{57.24}    & \textbf{62.08}       & \textbf{64.21}      \\ \midrule
\multicolumn{4}{l}{$\mathrm{ATLOP-DeBERTa}_{Large}$\cite{zhou2021document}}          & 62.16 & 64.01 & 68.45        & 59.63        & 62.12       & 64.08      \\
\multicolumn{4}{l}{$\mathrm{\textbf{FCDS-DeBERTa}}_{Large}$}           & \textbf{64.12} & \textbf{66.17} & \textbf{70.19}        & \textbf{58.73}        & \textbf{64.03}       & \textbf{65.86}    \\ \midrule \\
\end{tabular}
\end{table*}

The experimental results listed in Table \ref{results in docred} show that our model can achieve leading performance in DocRE data in the general domain. The proposed model outperforms the dependency graph-based methods SagDRE \cite{SagDRE} by margins of 2.29\% and 1.94\% on the test set in terms of Ign F1 and F1, respectively, indicating that the combination of dependency and constituency syntax is useful for document-level relation extraction. Our model improves the IgnF1 score on the test set by 1.56\% over the state-of-the-art method LARSON \cite{duan-etal-2022-just}, which uses the constituency tree to predict the evidence for DocRE and uses the constituency and dependency syntax separately. The advance confirms that information of different granularity can assist relation extraction in DocRE. \par

\subsection{Results on Biomedical Datasets}

\begin{table}[]
\centering
\caption{F1 Results (\%) of relation extraction on the test set of CDR and GDA.}
\label{CDR/GDA}
\resizebox{\linewidth}{!}{
\begin{tabular}{lcc}
\hline
Model          & CDR            & GDA            \\ \hline
LSR-BERT\cite{nan-etal-2020-reasoning}       & 64.80           & 82.20           \\
SciBERT\cite{zhou2021document}        & 65.10           & 82.50           \\
ATLOP-SciBERT\cite{zhou2021document}  & 69.40           & 83.90           \\
EIDER-SciBERT\cite{xie2022eider}  & 70.63          & 84.54          \\
SagDRE-SciBERT\cite{SagDRE} & 71.80          & -              \\ 
LARSON-SciBERT\cite{duan-etal-2022-just} & 71.59          & 86.02              \\ 
DREEAM-SciBERT\cite{ma-etal-2023-dreeam} & 71.55          & 84.51              \\ 
$\mathrm{\textbf{FCDS-SciBERT}}$           & \textbf{72.62} & \textbf{87.39} \\ \hline
\end{tabular}}
\end{table}

In addition to general domain DocRE methods, we also compare our model with various advanced methods including LSR \cite{nan-etal-2020-reasoning}, sciBERT \cite{zhou2021document}, ATLOP \cite{zhou2021document}, EIDER \cite{xie2022eider}, SagDRE \cite{SagDRE} and LARSON \cite{duan-etal-2022-just} on two biomedical domain datasets CDR and GDA. \par 
Experimental results are listed in Table \ref{CDR/GDA}. In summary, our model achieves significant improvements over two tested datasets (0.93\% on CDR and 1.27\% on GDA). GDA and CDR have more sentences than DocRED, thus our model can deal with complex documents, further demonstrating its generality.

Furthermore, to assess the significance of improvements, we perform a two-sample t-test comparing our approach with five other methods in three datasets. The obtained p-values are presented in Table \ref{p}. It can be seen that all values are less than 0.05, demonstrating a significant improvement of FCDS.

\begin{table}[]
\caption{Two-sample t-test on all datasets. Use BERT for embeddings on DocRED, and SciBERT on CDR and GDA.}
\label{p}
\resizebox{\linewidth}{!}{
\begin{tabular}{lcc}
\hline
Model\ \textbackslash \ Metric       & IgnF1            & F1            \\ \hline
LSR\cite{nan-etal-2020-reasoning}       & 0.0146           & 0.0129           \\
ATLOP\cite{zhou2021document}  & 0.0240           & 0.0213           \\
EIDER\cite{xie2022eider}  & 0.0337          & 0.0313          \\
LARSON\cite{duan-etal-2022-just} & 0.0352          & 0.0341              \\ 
DREEAM\cite{ma-etal-2023-dreeam} & 0.0470          & 0.0439              \\ \hline
\end{tabular}}
\end{table}

\begin{figure}[!t]
\centering{\includegraphics[width=\linewidth]{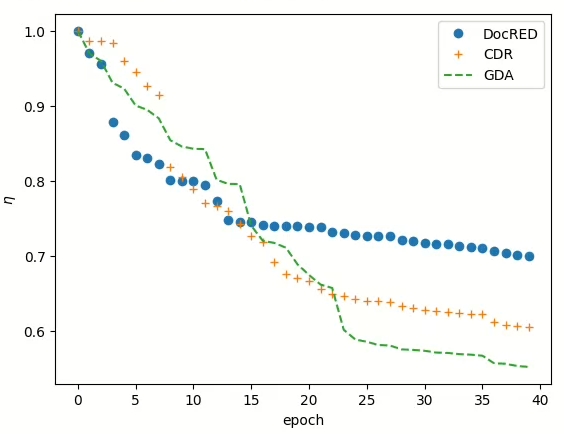}}
\caption{The learning curve of $\eta$ on DocRED, CDR and GDA datasets. }
\label{learning}
\end{figure}

\begin{table}[]
\caption{Ablation study of our model in dev set of DocRED.}
\label{ablation}
\begin{tabular}{llc}
\hline
Ablation              & \multicolumn{1}{c}{F1} & \multicolumn{1}{l}{IgnF1} \\ \hline
FCDS             & 64.42                  & 62.61                     \\ 
separate syntax   & 62.59                  & 60.63                     \\
w/o document node & 62.78                   & 60.92                     \\
w/o constituency tree & 62.15                  & 60.22                     \\ 
w/o dependency graph   & 61.63                  & 59.41                     \\
\hline
\end{tabular}
\end{table}

\begin{table*}
\resizebox{\textwidth}{!}{
\begin{tabular}{ccclcclcc}
\hline
         & \multicolumn{2}{c}{DocRED}             &  & \multicolumn{2}{c}{CDR}            &  & \multicolumn{2}{c}{GDA}              \\ \cline{2-3} \cline{5-6} \cline{8-9} 
         & with document  & w/o document &  & with document& w/o document &  & with document & w/o document  \\
         & node &  node &  &  node& node &  &  node &  node \\ \hline
avg \& std     & 6.19 $\pm$ 1.02           & 7.23 $\pm$ 1.55          &  & 5.96 $\pm$ 2.26            & 6.44 $\pm$ 2.23         &  & 6.82  $\pm$ 1.99           & 7.38 $\pm$ 2.43  \\
max & 7    & 9 &  & 8& 12 &  & 11            & 14  \\
min   & 4  & 4 &  & 4        & 4  &  & 3              & 4     \\
 \hline

\end{tabular}}
\label{tab:3}
\caption{The average/max/min length and standard deviation of entity distances with document node and w/o document node in DocRED, CDR, GDA.}\label{attr}
\end{table*}

\subsection{Ablation Study}
To exhaustively understand how each component of our method contributes to the final performance, we perform ablation studies to analyze the function of different syntaxes. In Figure \ref{learning}, we plot the learning curve of $\eta$ in Eq. \ref{eta}. We can observe that $\eta$ decreases from the initial value of 1.0, indicating that the dependency graph plays a more vital role during the process, possibly due to the fact that the dependency graph combines the features of the constituency tree and captures fine-granularity information. \par

Furthermore, we remove one component at a time and assess the resulting model using the dev set in DocRED in Table \ref{ablation}. For separate syntax case, we do not use the constituency tree to enhance the dependency graph and observe that F1 and IgnF1 decrease by 0.78\% and 0.94\%, indicating that fusing syntax is beneficial for DocRE. Then we remove the document node from the dependency graph, dependency graph, and constituency tree, respectively. For w/o document node, we observe that F1 and IgnF1 decrease by 0.59\% and 0.65\%. We speculate that the document node reduces the distance between entity pairs, which contributes to DocRE. For the w/o dependency graph, only the constituency tree is incorporated for relation prediction. We can observe that the F1 score decreases by 2. 16\%. Similar trends occur when we remove the constituency tree, where F1 decreases to 62.15\%. Therefore, both dependency graphs and constituency trees are significant for our models.

Finally, to examine the impact of document nodes on reducing the distance between entity pairs, we analyze the changes in entity distances before and after adding document-level nodes in three datasets in Table \ref{attr}. Through a random selection of 600 cases, we assess the average, maximum, minimum, and standard deviation of entity distances with and without document nodes. Our finding indicates a reduction in average entity distances, suggesting that document nodes could improve DocRE by simplifying interactions between entities.

\subsection{Case Study}

\begin{figure}[t]
\begin{subfigure}{1.0\columnwidth}
\includegraphics[width=\textwidth]{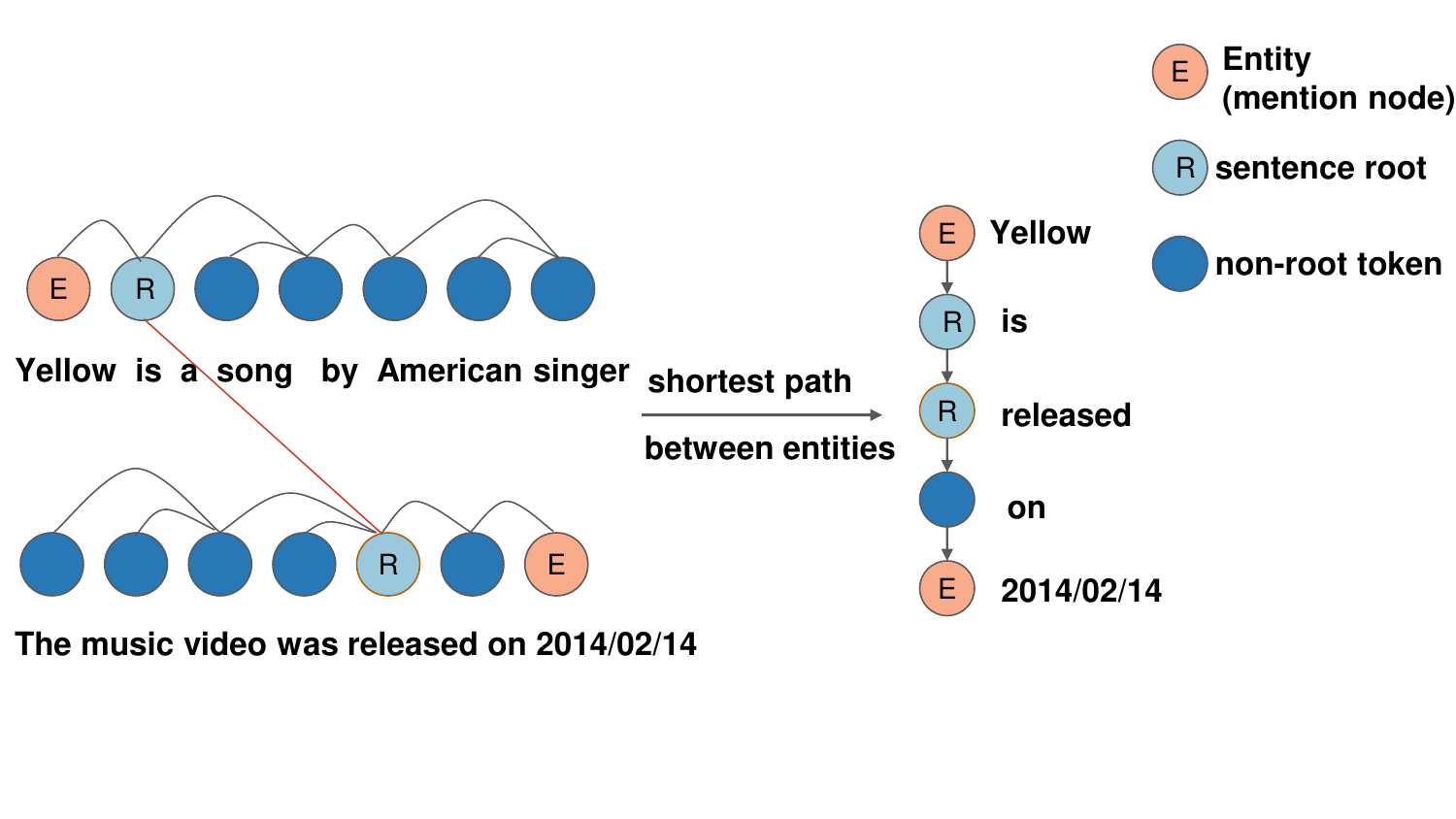}
\caption{ In this case, the dependency graph plays a positive role in DocRE by extracting the shortest path '\textbf{Yellow} is released on \textbf{2014/02/14}' between entities \textbf{Yellow} and \textbf{2014/02/14}.
} 
\label{fig_casea}
\vspace{0.5cm}
\end{subfigure}
\begin{subfigure}{1.0\columnwidth} 
\includegraphics[width=\textwidth]{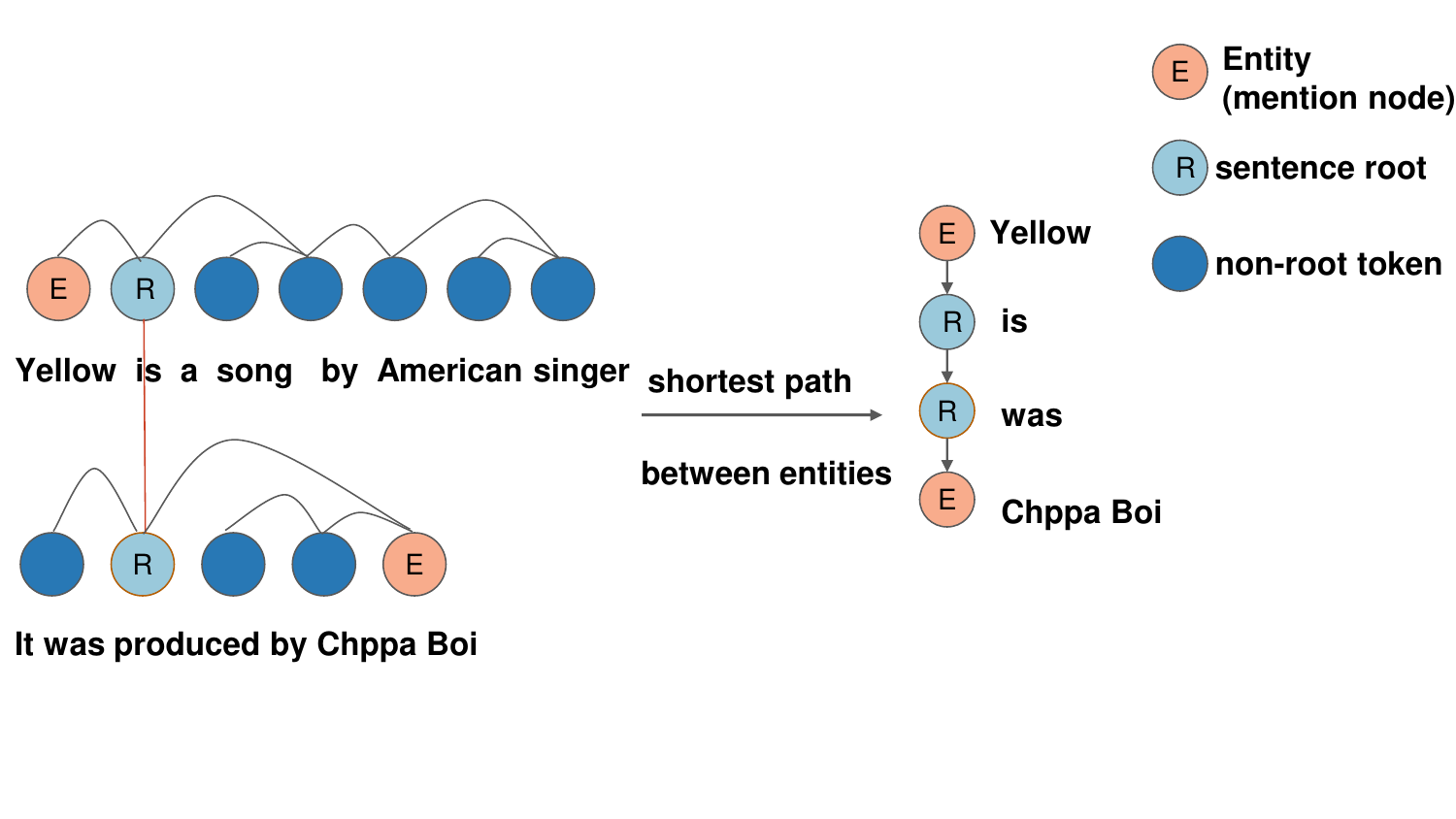}

\caption{In this case, the dependency graph plays an ambivalent role in DocRE. It extracts the shortest path '\textbf{Yellow} is was \textbf{Chppa Boi}' between \textbf{Yellow} and its author \textbf{Chppa Boi}. However, due to the inaccuracy of the dependency parser, the model cannot capture the keyword \textbf{produced}, therefore it is hard to predict the real relation between entities.}
\label{fig_caseb}
\end{subfigure}
\caption{Two different cases in DocRED.}\label{case}
\end{figure}

To better understand the bottleneck of FCDS and inspire future work, we conduct a case study to investigate the predictions that FCDS makes. The result is shown in Figure \ref{case}.\par

The first case in Figure \ref{fig_casea} is a successful dependency graph and illustrates how the dependency syntax helps the model complete DocRE. In this case, the dependency parser successfully parses the dependency syntax within two sentences and extracts vital keywords to manage the prediction. With the short path "Yellow is released on 2014/02/14.", simple sentence-level models can finish the prediction. \par

The second case in Figure \ref{fig_caseb} illustrates the motivation behind our methods, which is to alleviate the inaccuracy of the dependency parser and the failure of selecting keywords for entities. Although the dependency syntax is a useful tool for DocRE, it sometimes fails to identify the relevant keywords for entities. In this case, the shortest path "Yellow is was Chppa Boi" is far from finishing the prediction, while the real keyword "produced" is not selected in the path. Furthermore, errors in the dependency parser lead to a severe lack of information. To address this issue, we utilize the constituency tree and fuse the information from dependency and constituency syntax, which organizes different words of a single sentence hierarchically and can aggregate the sentence-level information naturally towards the sentence root. By doing so, the sentence root can obtain the mixed information of the root itself and the sentence information, adding the information of the path and alleviating the inaccuracy of the dependency parser. \par

\section{Conclusion}

In this work, we propose a novel model for the document-level relation extraction task. Our model exploits two types of extra syntax information, namely dependency syntax and constituency syntax. 
GCN and Tree-LSTM are adopted to encode the two types of information. Furthermore, by using the constituency tree to enhance the dependency graph and adding a document node in the dependency graph, we can improve the expression capability of the dependency graph and better capture long-distance correlations. Experiments on three public DocRE datasets demonstrate that our model outperforms the existing method. In the future, we plan to select the most conducive sentences for entity pair by constituency tree, which captures the information from another perspective and complements the dependency graph.

\section{Acknowledgements}

This work was supported by the National Natural Science Foundation of China (Nos. 62276053 and 62273071) and High-performance Computing Platform of UESTC.

\section{Bibliographical References}

\bibliographystyle{lrec-coling2024-natbib}
\bibliography{lrec-coling2024-example}

\bibliographystylelanguageresource{lrec-coling2024-natbib}
\bibliographylanguageresource{languageresource}

\end{document}